\documentclass[sigconf]{acmart}
\usepackage{algorithmic}
\usepackage{graphicx}
\usepackage{textcomp}
\usepackage{xcolor}
\usepackage{multirow}
\usepackage{multicol}
\usepackage{caption}
\usepackage{hyperref}
\usepackage{enumitem}
\usepackage[most]{tcolorbox}

\AtBeginDocument{%
  }

\setcopyright{acmlicensed}
\copyrightyear{2025}
\acmYear{2025}
\acmDOI{XXXXXXX.XXXXXXX}
\acmConference[BIOKDD'25]{24th International Workshop on Data Mining in Bioinformatics}{August 03--07,
  2025}{Ontario, Canada}
\acmISBN{978-1-4503-XXXX-X/2025/06}

\begin{document}

\title{Interpreting Biomedical VLMs on High-Imbalance Out-of-Distributions: An Insight into BiomedCLIP on Radiology}


\author{Nafiz Sadman and Farhana Zulkernine}
\affiliation{%
  \institution{School of Computing}
  \city{Queen's University}
  \country{Canada}
}
\email{ {sadman.n,farhana.zulkernine}@queensu.ca }

\author{Benjamin Kwan}
\affiliation{%
  \institution{Department of Diagnostic Radiology}
  \city{Kingston Health Sciences Centre}
  \country{Canada}
}
\email{benjamin.kwan@kingstonhsc.ca}

\begin{abstract} 
  In this paper, we construct two research objectives: i) explore the learned embedding space of BiomedCLIP, an open-source large vision language model, to analyse meaningful class separations, and ii) quantify the limitations of BiomedCLIP when applied to a highly imbalanced, out-of-distribution multi-label medical dataset. We experiment on IU-xray dataset, which exhibits the aforementioned criteria, and evaluate BiomedCLIP in classifying images (radiographs) in three contexts: \textit{zero-shot} inference, \textit{full finetuning}, and \textit{linear probing}. The results show that the model under zero-shot settings over-predicts all labels, leading to poor precision and inter-class separability. Full fine-tuning improves classification of distinct diseases, while linear probing detects overlapping features. We demonstrate visual understanding of the model using Grad-CAM heatmaps and compare with 15 annotations by a radiologist. We highlight the need for careful adaptations of the models to foster reliability and applicability in a real-world setting. The code for the experiments in this work is available and maintained on \href{https://github.com/Nafiz95/BioVLM_Eval_CXR}{\color{blue}{GitHub}}.

\end{abstract}

\ccsdesc[300]{Computing methodologies~Visual inspection}
\ccsdesc[500]{Information systems~Retrieval effectiveness}
\ccsdesc[500]{Information systems~Clustering and classification}
\ccsdesc[300]{Human-centered computing~Visualization design and evaluation methods}
\keywords{Biomedical vision-language models, Visualizations, Evaluations, Quantitative analysis, Qualitative Analysis, Radiology }
\begin{teaserfigure}
  \includegraphics[width=\textwidth,height=3in]{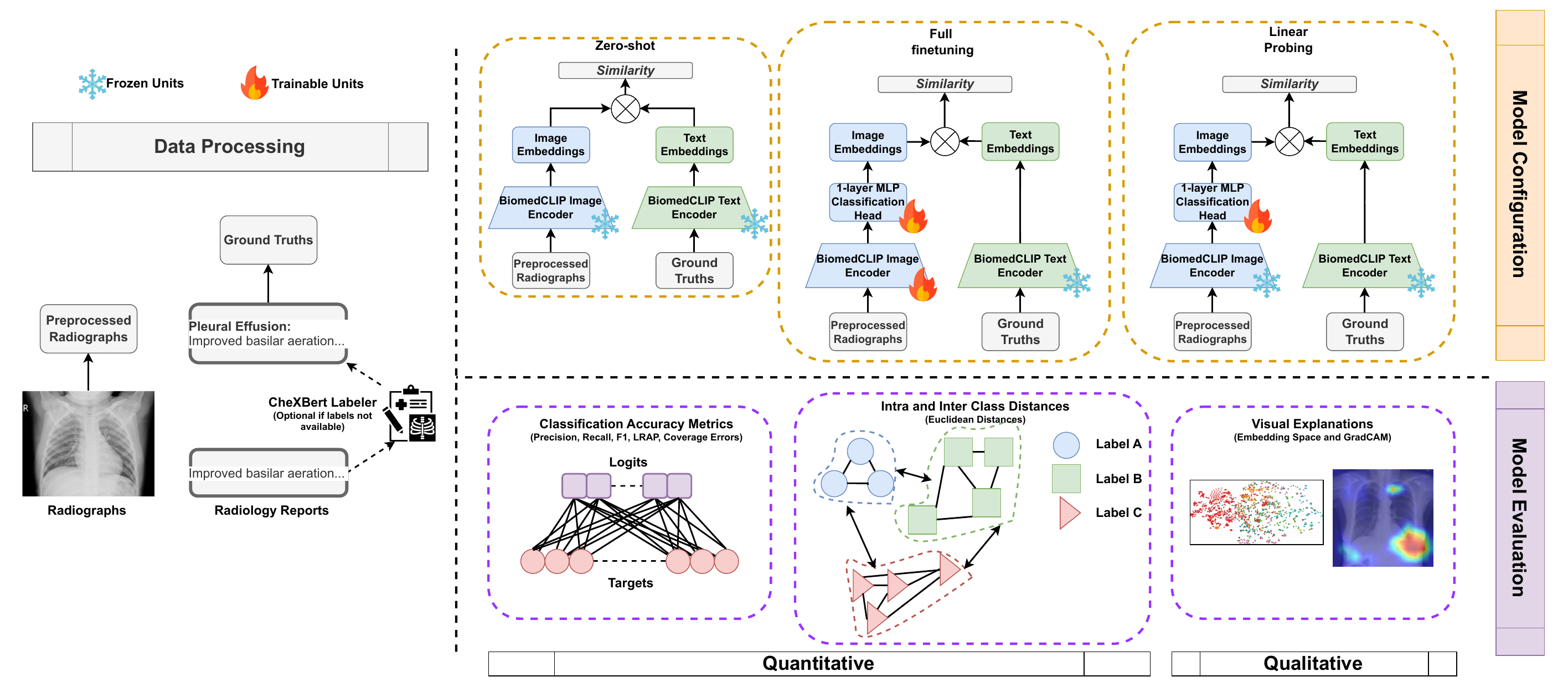}
  \caption{Quantitative and qualitative evaluation framework for BiomedCLIP under three settings: zero-shot, full finetuning, and linear probing.}
  \Description{Evaluation Framework for BiomedCLIP}
  \label{fig:teaser}
\end{teaserfigure}

\received{20 February 2007}
\received[revised]{12 March 2009}
\received[accepted]{5 June 2009}

\maketitle
\section{Introduction}
Artificial intelligence-assisted radiology informatics (AIRI) remains challenging to deploy due to the complexity of radiographs (e.g., X-rays with non-distinctive features among overlapping diseases) and the subjective nature of radiologists’ reports \cite{jing2017automatic}. Some examples of AIRI include image segmentation \cite{10643318}, image classification \cite{atmakuru2024deep}, and report generation from radiographs \cite{10545538}. To process radiographs, early work used convolutional neural network (CNN) variants, like CNN-RNN, to extract image features \cite{shin2016deep}. Vision transformers (ViTs), introduced by Dosovitskiy et al.\ \cite{dosovitskiy2020image}, are more popular and can capture global features more effectively than CNNs. Meanwhile, radiology reports are processed by language models like BERT \cite{smit2020chexbert} to extract semantic text features and classify reports into disease categories. However, many studies have demonstrated that combining image and textual features in a contrastive alignment outperforms unimodal approaches \cite{Boecking,Bannur_2023_CVPR,biomedclip}. Thus, vision-language models (VLMs) have been proposed.
One limitation inherent to the contrastive nature of VLMs is that they require large amounts of training data to learn effective image–text pair representations \cite{clip}. Although VLMs excel at tasks involving distinct and well-separated multi-class classification, \textbf{\textit{their performance can degrade when classifying limited data with complex or closely related classes}} (e.g., distinguishing ‘Pneumonia’ from ‘Consolidation’ in biomedical datasets) \cite{Bannur_2023_CVPR,varoquaux2022machine}. Biomedical datasets exhibit complex relationships, multi-label dependencies, and extreme class imbalance. Moreover, rare diseases remain underrepresented, leading to low detection by automated systems, i.e., machine learning models \cite{wojtara2023artificial}. To address this issue, researchers have proposed domain-specific pretraining of VLMs—training them on tailored, domain-specific datasets—to enhance performance on such tasks \cite{chambon2022adapting}. However, this specialized pretraining \textbf{\textit{may compromise domain generalization}}, as models optimized for a particular domain might perform less effectively on out-of-distribution (OOD) data \cite{zech2018variable}. Therefore, we pose the following research question:
\begin{quote}
    \textit{\textbf{Can large, pretrained vision–language models accurately classify images in a multi‐label, imbalanced, out‐of‐distribution biomedical dataset?}}
\end{quote}

\noindent Based on this, we define two research objectives:
\begin{enumerate}
    \item To quantitatively analyze the inter‐ and intra‐class distances in the learned embeddings of the vision–language model.  
    \item To evaluate the model’s performance limitations on a highly imbalanced, out‐of‐distribution, multi‐label medical dataset using a multi‐faceted set of performance metrics.
\end{enumerate}
    
\noindent We hypothesize that:

\begin{quote}
    \textbf{\textit{Although zero‐shot inference with large pretrained vision–language models provides a strong baseline on multi‐label, imbalanced, out‐of‐distribution biomedical datasets, moderate adaptation strategies (e.g.\ linear probing) will yield further performance gains at a reasonable computational cost—substantially lower than that required for full end‐to‐end fine‐tuning}}
\end{quote}



To justify our point, we experiment with BiomedCLIP \cite{biomedclip}, an open-source VLM trained on 15 million medical image–text pairs (the PMC-15M dataset). To date, BiomedCLIP outperforms the radiology-specific BioViL model \cite{boecking2022most} on the RSNA pneumonia detection benchmark and achieves a mean accuracy of 75.5\% across five zero-shot classification datasets, a 12-point improvement over general-domain CLIP. It also achieves a top-1 recall of 56\% in cross-modal retrieval over the PMC-15M test set. Finally, on medical VQA (VQA-RAD), it attains 75.2\% accuracy—surpassing the former best of 68.5\%—further confirming its broad, state-of-the-art performance across classification, retrieval, and reasoning tasks.

We demonstrate the overall workflow in Fig. \ref{fig:teaser}. We evaluate BiomedCLIP on the IU-xray dataset \cite{iuxray}, a 14-label multi-class benchmark that is highly imbalanced (2,400 “No Finding” vs. 20 “Pneumothorax” samples). Morever, BiomedCLIP is not pretrained on this dataset, which renders it as OOD. We assess its performance under three model adaptations: zero-shot inference, full fine-tuning, and linear probing. These three adaptation span a continuum of computational cost and performance trade‐offs \cite{Goyal_2023_CVPR}. Zero-shot is the go-to method that does not require in-domain training due to its massive pretrained knowledge representations. It is computationally less expensive than full fine-tuning, which requires more computational resources (e.g.\ GPU memory), to retrain the weights of the entire network \cite{pmlr-v139-radford21a}. Conversely, linear probing freezes the encoder and trains only a lightweight classification head, offering a low‐compute adaptation that often yields substantial accuracy gains while preserving the quality of the pretrained representations by the pretrained model. 

For each settings (or adaptations), we compute per-class precision, recall, and F1, as well as overall multi-label metrics (macro-F1, exact-match accuracy, LRAP, coverage error, ranking loss). We also quantify embedding-space separability via the inter/intra-class Euclidean-distance ratio, and we visually inspect its explanations via Grad-CAM \cite{Selvaraju_2017_ICCV}.
These evaluation metrics are standard for assessing both detection quality and ranking performance in multi-label classification \cite{6471714}. Additionally, we obtained 15 radiologist-annotated radiographs, enabling direct visual comparison with our Grad-CAM heatmaps. This represents a crucial step toward validating model interpretability in real-world clinical settings.





We observe two notable findings from our experiments:
\begin{enumerate}
    \item Full fine-tuning exhibits a higher inter-/intra-class distance ratio than zero-shot inference and linear probing, which is counterintuitive \---\ one would normally expect end-to-end tuning to yield superior class separability. Interestingly, linear probing achieves a comparable ratio to zero-shot inference.
    
    \item Zero-shot BiomedCLIP produces significant false positives and low precision for rare diseases (i.e.,\ rare disease. While full fine-tuning improves classification of well-represented diseases, linear probing enhances detection of rare-class features; notably, its overall performance is on par with that of full fine-tuning.
\end{enumerate}


The paper is organized as follows: we present recent related literature in Section \ref{rel}. Then, we describe the data properties of the dataset used in this research in Section \ref{data_desc}, and elaborate our methodology to conduct the experiments in Section \ref{method}. We discuss the results and findings in Section \ref{results}. Finally, we share our thoughts and future directions in Section \ref{concl}.

\section{Related Work}
\label{rel}

\noindent\textbf{Vision–Language Models (VLMs) in Biomedicine.}
Recent foundation models such as CLIP showed that aligning images and text in a shared embedding space can yield remarkable performance on vision tasks without task-specific training, enabling capabilities like zero-shot image classification and cross-modal retrieval.  Researchers have extended VLMs to specialized domains such as biomedicine, where datasets are multimodal but often limited and lack ground-truth labels \cite{koleilat2024biomedcoop}.

Efforts to adapt VLMs to biomedical data have focused on self-supervised learning from medical images and associated text, such as radiology reports.  For example, ConVIRT \cite{pmlr-v182-zhang22a} trained dual image/text encoders on paired chest X-rays and radiology reports using bidirectional contrastive learning, achieving improved transfer learning for medical image classification.  GLoRIA \cite{Huang_2021_ICCV} introduced a global–local alignment mechanism: in addition to matching whole-image and report embeddings, it uses cross-attention to link image sub-regions with semantic phrases in the report, thereby capturing pathology-specific visual details and improving interpretability.  These domain-specific pretraining approaches demonstrated that medical image representations benefit from joint text supervision, yielding higher downstream accuracy than vision-only counterparts \cite{wang2022medclip}.

Large-scale biomedical VLMs such as BioViL \cite{Bannur_2023_CVPR} combined a radiology-tuned language encoder with a vision backbone in a contrastive framework, using millions of hospital image–report pairs.  BioViL achieved state-of-the-art results on multiple chest X-ray benchmarks (e.g., abnormality classification and natural-language inference) by tailoring the text encoder to clinical language.  Similarly, MedCLIP \cite{wang2022medclip} leveraged unpaired medical images and text: it decoupled image–text corpora to create synthetic pairings and introduced a semantic matching loss to handle noisy correspondences.  MedCLIP proved remarkably data-efficient—using only 10\% of the usual pretraining samples, it surpassed prior radiology VLMs like GLoRIA on zero-shot label prediction and supervised classification.

These biomedical VLMs have been evaluated on a wide range of medical imaging tasks, often outperforming conventional methods.  For image classification, BiomedCLIP set new state-of-the-art results on standard radiology tasks.  While these models demonstrate the feasibility of multimodal diagnostic reasoning, a sizable gap remains between their performance and that of human radiologists \cite{zambrano2025clinically}, underscoring the need for domain-specialized VLMs and careful evaluation on clinically realistic tasks.

\vspace{0.5em}
\noindent\textbf{Adaptation Strategies: Zero-Shot, Fine-Tuning, and Linear Probing.}
A crucial question in applying VLMs to biomedical tasks is how best to adapt the pretrained model to target data.  At one extreme, VLMs can be used in a zero-shot manner—treating each disease as a text prompt (e.g., ``X-ray showing pneumonia'') and selecting the label whose embedding best matches the image embedding \cite{wang2022medclip}.  However, zero-shot accuracy often lags behind supervised methods, especially for subtle or rare findings, due to phrasing mismatches and limited exposure to specific visual manifestations.

At the other extreme, full fine-tuning on in-domain labels usually yields the highest accuracy, as shown by BiomedCLIP and other models fine-tuned for chest X-ray classification or segmentation \cite{zhang2023knowledge}.  Yet, full fine-tuning of a large multimodal model is computationally expensive and risks overfitting when datasets are small.  As a middle ground, linear probing—training only a lightweight classifier on frozen image embeddings—has emerged as an efficient adaptation: it often recovers much of the performance gap to full fine-tuning at a fraction of the computational cost \cite{schafer2024overcoming}.  Overall, the consensus is that naive zero-shot inference is suboptimal in medicine, and minimal task-specific adaptation—via fine-tuning, linear probing, or learned prompts—is typically required to capture the fine-grained, domain-specific nuances of biomedical datasets.

\vspace{0.5em}
\noindent\textbf{Limitations of Existing Biomedical VLMs.}
Despite rapid progress, biomedical VLMs still face key challenges in clinical deployment.  One major issue is domain shift: models trained and evaluated on similar datasets (e.g., MIMIC-CXR) can degrade markedly when confronted with out-of-distribution data from different hospitals, patient populations, or imaging modalities.  

Another limitation is interpretability.  In high-stakes medical contexts, clinicians must understand why a model made a given prediction.  Most VLMs function as black boxes, with limited built-in explainability.  Some methods, such as GLoRIA, provide attention maps linking words to image regions, and others employ Grad-CAM post-hoc to highlight salient areas.  However, recent audits indicate that these saliency maps are often misaligned with true pathology locations \cite{mcinerney2022s}, which can undermine clinician trust.

\vspace{0.5em}
\noindent\textbf{Our Contributions.}
In light of these gaps, our work pushes the boundary on evaluating and interpreting a biomedical VLM under challenging conditions.  Whereas prior studies report only aggregate performance, we conduct a fine-grained analysis on a highly imbalanced, out-of-distribution radiography dataset (IU-Xray) to assess how BiomedCLIP handles both common and rare findings.  We compare adaptation regimes—zero-shot, linear probing, and full fine-tuning—and reveal nuances such as linear classifiers outperforming end-to-end tuning on mid-frequency diseases.  Unlike earlier work that treats VLMs as black boxes, we integrate embedding-space analysis and radiologist-validated Grad-CAM heatmaps to deliver a more transparent evaluation, helping bridge the gap between bench-top performance and trustworthy clinical deployment.

\section{Dataset Description}
\label{data_desc}
\begin{figure}[!h]
    \centering
    \includegraphics[scale=0.32]{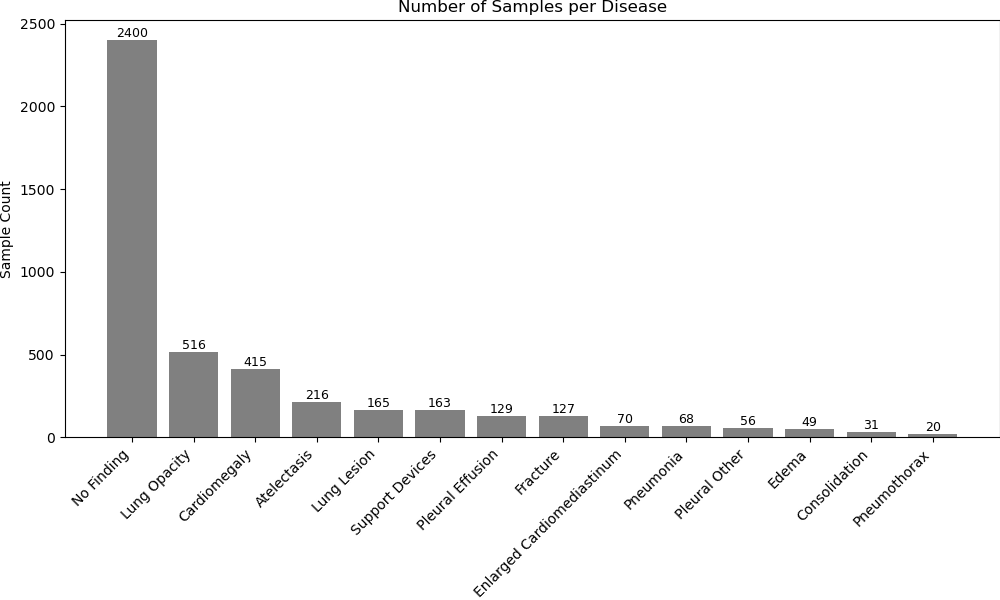}
    \caption{Data sample distribution of 14 disease classes of IU-xray dataset.}
    \Description{Data sample distribution of 14 disease classes of IU-xray dataset.}
    \label{fig:treemap}
\end{figure}

We evaluate BiomedCLIP on the IU-xray dataset developed by Indiana University \cite{iuxray} in 2017. The dataset has 7,470 radiographs (frontal and lateral) of the chest and 3,955 reports of 3,851 patients from two large hospital systems within the Indiana Network for Patient Care database. All identifiable information of patients is anonymized. IU-xray dataset is one of the benchmarks used in several radiology report analysis and generation tasks. More importantly, it is one of the smaller datasets which is often not used for training VLMs.

The dataset is unlabeled. Therefore, we apply Standford's pre-trained BERT model, CheXbert \cite{smit2020chexbert}, to label each report as one or more of the 14 chest disease categories. CheXbert outperforms previous labelers for chest radiographs with 80\% validated accuracy. The categories extracted (refer to the labels in Fig. \ref{fig:treemap}) from CheXbert are: \textit{Enlarged Cardiomediastinum}, \textit{Cardiomegaly},\textit{ Lung Opacity},\textit{ Lung Lesion}, \textit{Edema}, \textit{Consolidation}, \textit{Pneumonia}, \textit{Atelectasis}, \textit{Pneumothorax}, \textit{Pleural Effusion}, \textit{Pleural Other}, \textit{Fracture}, \textit{Support Devices}, and \textit{No Finding}. Fig. \ref{fig:treemap} demonstrates the highly imbalanced class distribution in the dataset, with `\textit{No Finding}' being the majority class and `\textit{Pneumothorax}' being the minority class.

Additionally, to quantify the model's classification on broader categories, we assign the following domain specific labels (verified correctness by a radiologist): i) Cardiovascular \---\ \textit{Enlarged Cardiomediastinum}, \textit{Cardiomegaly}, ii) Skeletal \---\ \textit{Fracture}, iii) Device \---\ \textit{Support Devices}, and iii) Pulmonary \---\ \textit{Pneumonia}, \textit{Consolidation}, \textit{Atelectasis}, \textit{Pneumothorax}, \textit{Pleural Other}, \textit{Pleural Effusion}, \textit{Edema}, \textit{Lung Opacity}, \textit{Lung Lesion}.

\section{Methodology}
\label{method}
There are three components to implement and evaluate BiomedCLIP, as presented in Fig. \ref{fig:teaser}. We denote radiograph ($r_g$), CheXbert labeled reports ($r_r$), and associated diseases($r_t$) as triplets in the IU-Xray dataset as $(r_{g1},r_{r1},r_{t1}),(r_{g2},r_{r2},r_{t2}),..(r_{gn},r_{rn}, r_{tn}) \in D$, where \textit{n} is the total number of data points in the IU-Xray dataset, $D$, and $r_t \subseteq Diseases$. We also denote the BiomedCLIP image encoder as $B_I$ and the text encoder as $B_T$. Furthermore, for linear probing, we add one fully connected multiperceptron classification head, $H_T$, to the last layer of $B_I$. 

\subsection{Data Processing}
Each radiograph, $r_{gk} \in D_{k}$, is first resized to 224x224 pixels with center cropping, followed by mean-normalization of pixel values. Each label of the corresponding reports, $r_{rk} \in D_{k}$,  is tokenized using PubMedBERT tokenizer \cite{pubmedbert} and padded to 256 tokens.  These preprocessing steps are aligned with BiomedCLIP as outlined by Zhang et al.\ \cite{biomedclip}.

\subsection{Model Settings}
The preprocessed radiographs, $r_{gk}$, and tokenized labels $r_{rk}$ are used in three model settings: \emph{zero-shot}, \emph{fine-tuning}, and \emph{linear probing}.
\begin{enumerate}
    \item \textbf{Zero-shot}: Each radiograph $r_{gk}$ is processed by $B_I$, while its corresponding label $r_{rk}$ is processed by $B_T$. This yields contextual image embeddings $E(r_{gk})$ and text embeddings $E(r_{rk})$ for each $k \in \{1,2,...,n\}$.

    \item \textbf{Fine-tuning}:  
    We freeze \(B_I\) and train only the new head \(H_T\) for one \emph{warm-up epoch}. We use the binary cross-entropy (BCE) loss and the AdamW optimizer with weight decay \(10^{-2}\) to stabilize the random head initialization.  After the warm-up, we \emph{unfreeze the entire} visual encoder \(B_I\) and continue to jointly optimize all parameters of \(B_I\) and \(H_T\).  We employ a cosine-annealing learning rate (LR) schedule, dropping the base LR by a factor of 10 when unfreezing, and apply early stopping based on the validation BCE loss to prevent overfitting.

  \item \textbf{Linear probing}:  
    We keep \(B_I\) fully frozen for the entire training run and train only \(H_T\) from scratch using BCE loss, AdamW, and a cosine LR schedule.  Early stopping is again governed by the validation BCE loss.
\end{enumerate}

\subsection{Experiment Setup}
To reiterate our objectives:
\begin{enumerate}
  \item Quantitatively and qualitatively evaluate BiomedCLIP's classification performance on the imbalanced OOD dataset (i.e., Iu-xray).

  \item Validate linear probing \---\ an alternative to fine-tuning \---\ for its classification performance and explainability.
\end{enumerate}

\noindent Therefore, we design three experiments:

\begin{enumerate}
  \item \textbf{Embedding‐space analysis.}  
    We compute inter‐ and intra‐class distances (and their ratio) on the learned image embeddings to quantitatively assess how well the model separates different disease categories.

  \item \textbf{Performance evaluation.}  
    We evaluate radiograph classification under zero‐shot, full fine‐tuning, and linear probing using the multi-label metrics (macro-F1, exact-match, LRAP, coverage error, label-ranking loss).

  \item \textbf{Qualitative attention inspection.}  
    We generate Grad-CAM visualizations for a random subset of test images in each setting and compare those heatmaps against radiologist annotations to understand what the model is focusing on and its consistency. We also inspect how these visualizations change if we extract the representations from earlier layers.
\end{enumerate}

\noindent
\textbf{Implementation details:}  
All models use the same train/val/test split (70/10/20), a batch size of 24, AdamW optimizer (weight decay 1e-2), and early stopping on validation BCE loss.  Image inputs are resized to 224×224 and normalized with BiomedCLIP’s mean/std, and text labels are tokenized and padded to 256 tokens with PubMedBERT.  

\subsection{Evaluation Metrics}

Across all three settings \---\ zero‐shot, full fine‐tuning, and linear probing \---\ we compute the following quantitative metrics:

\noindent \textbf{\textit{Embedding‐space separability}:} We report inter‐class vs.\ intra‐class mean Euclidean distance of the disease classes, and their ratio in Table \ref{tab:overall_metrics}.

\noindent \textbf{\textit{Multi‐label classification metrics}:} We report per‐class F1 scores in Table \ref{tab:per_label_f1}. We report exact‐match accuracy (fraction of samples where the entire predicted label set exactly matches the ground truth), Label Ranking Average Precision (LRAP), Coverage error (number of top‐predicted labels needed to cover all true labels), and Macro-F1 scores in Table \ref{tab:overall_metrics}.

\noindent \textbf{\textit{Domain‐level metrics}:} We report Inter‐class vs.\ intra‐class mean Euclidean distance of the disease classes, and their ratio in Table \ref{tab:overall_metrics}.




      


\section{Results and Discussion}
\label{results}
\begin{figure*}
    \centering
    \includegraphics[width=\linewidth]{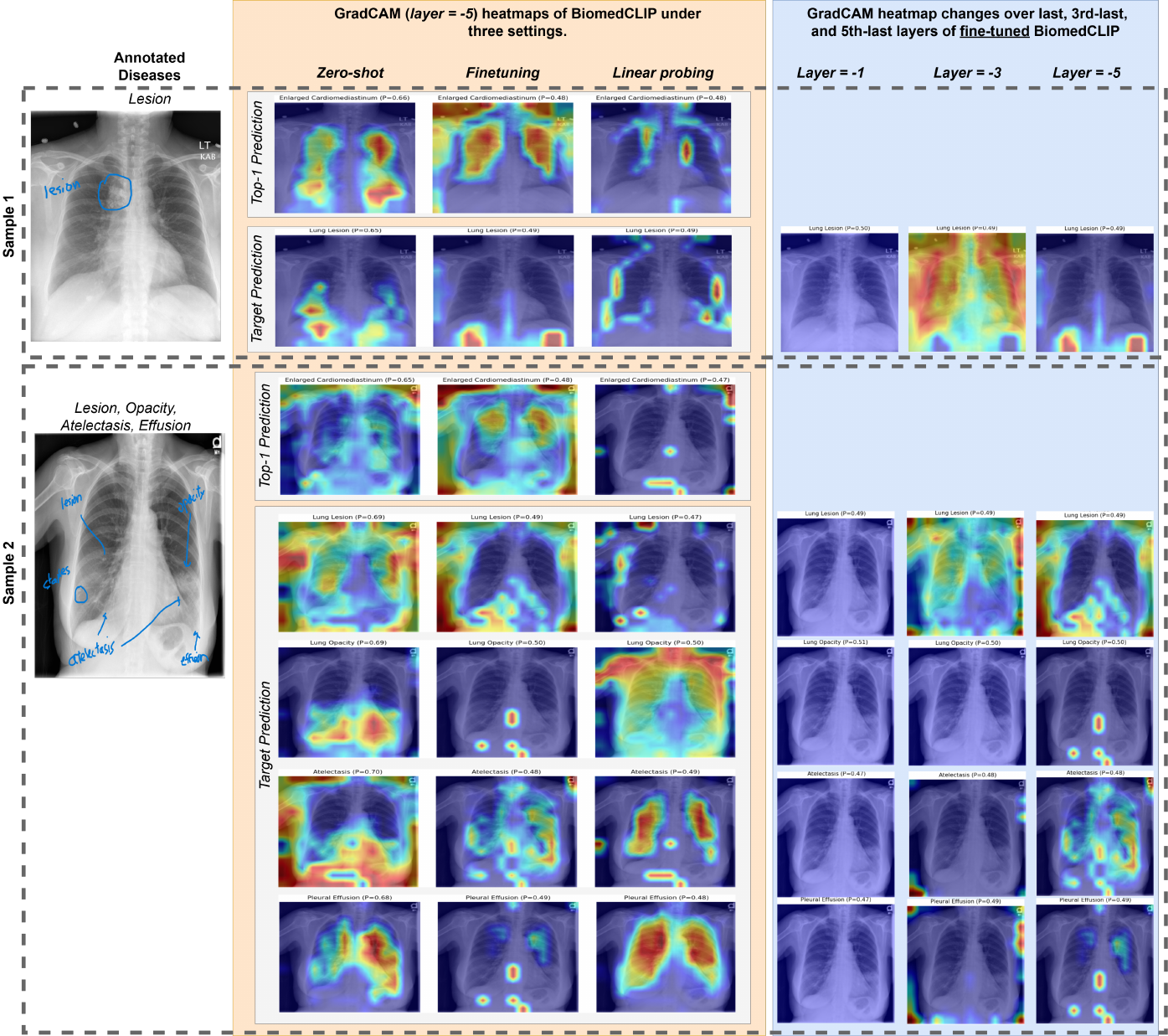}
    \caption{Grad-CAM visualizations of BiomedCLIP under zero-shot, full fine-tuning, and linear probing adaptations compared with radiologist-annotated ground-truth regions (blue). We also compare fine-tuned BiomedCLIP’s Grad-CAM outputs when using three different visual encoder depths.}
    \Description{Grad-CAM visualizations of BiomedCLIP under three evaluation settings \---\ zero-shot, full fine-tuning, and linear probing \---\ compared with radiologist-annotated ground-truth regions (blue). On the right, we also compare fine-tuned BiomedCLIP’s Grad-CAM outputs when using three different visual encoder depths.}
    \label{fig:gradcamsample}
\end{figure*}
\begin{table*}[ht]
\centering
\caption{Overall evaluation metrics for BiomedCLIP under three settings on test set. `zs', `ft', and `lp' represent zero-shot, fine-tuning, and linear probing, respectively. We highlight the best performance in blue and the worst in red.}
\label{tab:overall_metrics}
\begin{tabular}{lcccccccc}
\toprule
\textbf{Setting} & \textbf{Inter-class dist.$\uparrow$} & \textbf{Intra-class dist.$\downarrow$} & \textbf{Ratio $\uparrow$} & \textbf{F1 $\uparrow$} & \textbf{Exact match$\uparrow$} & \textbf{LRAP$\uparrow$} & \textbf{Coverage error$\downarrow$} & \textbf{Time (min)$\downarrow$} \\
\midrule
\textit{zs}         & 31.890 & \textcolor{red}{21.160} & \textcolor{red}{1.506} & \textcolor{red}{0.105} & \textcolor{red}{0.000} & \textcolor{red}{0.250} & \textcolor{red}{7.70} & \---\   \\
\textit{ft}  & \textcolor{red}{22.754} & \textcolor{blue}{12.756} & \textcolor{blue}{1.784} & \textcolor{blue}{0.235} & 0.134 & \textcolor{blue}{0.779} & \textcolor{blue}{2.750} & \textcolor{red}{15.47}  \\
\textit{lp}    & \textcolor{blue}{31.894} & 21.160 & 1.507 & 0.183 & \textcolor{blue}{0.143} & 0.741 & 3.077 & 6.10 \\
\bottomrule
\end{tabular}
\end{table*}

In this section, we present a detailed analysis of our three experiments: (1) embedding‐space analysis, (2) classification performance evaluation, and (3) We additionally comment on qualitative Grad‐CAM visualizations. 

\subsection{Experiment 1: Embedding space analysis}
\begin{tcolorbox}[
    colback=blue!10!white,    
    colframe=blue!60!black,   
    boxrule=0.8pt,            
    arc=3pt,                  
    left=6pt, right=6pt,      
    top=3pt, bottom=3pt       
]
\textbf{Key takeaways:}
\begin{itemize}[leftmargin=*]
  \item Full fine-tuning is the preferred strategy to achieve maximum class discriminability. 
  
  \item Linear probing achieves similar or better results than zero-shot with much less computational resources than full fine-tuning. These demonstrate that even without altering the core visual representations, linear probing recovers most of the performance achieved by full fine‐tuning. 
\end{itemize}
\end{tcolorbox}

\noindent \textbf{Cluster separation.} Fine‐tuning shrinks (reduces) intra‐class variance (21.16$\rightarrow$12.76) more than it shrinks inter‐class distances, enhancing the separation ratio from 1.51 to 1.78. This indicates that training the entire visual encoder encourages more dense and distinct class clusters. In contrast, linear probing leaves the backbone unchanged (inter$\approx$31.89, intra$\approx$21.17).

\noindent \textbf{Global classification metrics.}  By reshaping the embedding space, full fine‐tuning doubles the macro‐F$_1$ (0.105$\rightarrow$0.235), achieves nontrivial exact‐match accuracy (13.4 \%), and dramatically increases Label Ranking AP (0.250$\rightarrow$0.779) while reducing coverage error (7.70$\rightarrow$2.75). Importantly, linear probing, which simply trains a lightweight classification head on top of the frozen BiomedCLIP vision encoder, captures the majority of these gains at a fraction of the compute cost. Its inter/intra ratio (1.51) and coverage error (3.08) remain nearly identical to zero-shot, yet it still increases macro-F1 to 0.183, exact-match to 4.3\%, and LRAP to 0.741. Additionally, we record the training time for fine-tuning and linear probing, where the latter takes less than half the time of the former (15.47 mins$\rightarrow$6.10 mins).

\subsection{Experiment 2: Classification performance}
\begin{tcolorbox}[
    colback=blue!10!white,    
    colframe=blue!60!black,   
    boxrule=0.8pt,            
    arc=3pt,                  
    left=6pt, right=6pt,      
    top=3pt, bottom=3pt       
]
\textbf{Key takeaways:}
\begin{itemize}[leftmargin=*]
  \item Full fine‐tuning yields the highest overall F1 scores on abundant diseases. 
  
  \item Linear probing substantially outperforms zero‐shot inference, closing most of the gap to full fine‐tuning

  \item Both adaptation strategies struggle on extremely scarce diseases (e.g., Pneumothorax, Consolidation, Edema), but linear probing tends to generalize better on mid‐scarce diseases.
\end{itemize}
\end{tcolorbox}

Tables \ref{tab:per_label_f1} and \ref{tab:per_domain_f1} report F1 scores for each disease and for each disease domain, respectively. Full fine‐tuning achieves the highest absolute performance across almost all pathologies and domains; however, linear probing consistently outperforms zero‐shot inference and yields results comparable to full fine‐tuning. We categorize our observations according to disease prevalence:

\noindent \textbf{Abundant classes:} \textit{Classes with relatively more data samples (e.g., `No Finding` with 2400 samples, `Lung Opacity` with 516, and `Cardiomegaly` with 415).}
    Fine-tuning achieves the highest F1 (No Finding 0.803, Lung Opacity 0.354, Cardiomegaly 0.490), while linear probing comes close (No Finding 0.788, Lung Opacity 0.177, Cardiomegaly 0.421).  The gap narrows for `No Finding', suggesting that even a frozen encoder with a retrained classification head can achieve similar performance as fine‐tuning on very frequent labels.

\noindent \textbf{Rare classes:} \textit{Classes with relatively scarce data samples (e.g., `Pneumothorax` with 20, `Consolidation` with 31, and `Edema' with 49).} 
 Both fine-tuning and linear probing struggle when only a small set of samples exists (all F1$\approx$0 for Pneumothorax, Consolidation, Edema).  However, linear probing slightly outperforms fine-tuning on some mid‐frequency pathologies such as Pneumonia (F1 of 0.286 compared to fine-tuning F1 of 0.267), Atelectasis (F1 of 0.410 compared to fine-tuning F1 of 0.091), and Fracture (F1 of 0.087 compared to fine-tuning F1 of 0.000).  This suggests that linear probing can generalize better on classes with moderate but not extreme scarcity, perhaps by avoiding overfitting the small fine‐tuning set.

\begin{table}[ht]
\centering
\caption{Per‐label F1 scores for BiomedCLIP under three settings. `zs', `ft', and `lp' represent zero-shot, fine-tuning, and linear probing, respectively. We highlight the best performance in blue and the worst in red.}
\label{tab:per_label_f1}
\begin{tabular}{lccc}
\toprule
\textbf{Disease} & \textbf{\textit{zs}} & \textbf{\textit{ft}} & \textbf{\textit{lp}} \\
\midrule
Enlarged Cardiomediastinum & \textcolor{blue}{0.039} & \textcolor{red}{0.000} & 0.015 \\
Cardiomegaly               & \textcolor{red}{0.216} & \textcolor{blue}{0.490} & 0.421 \\
Lung Opacity               & 0.314 & \textcolor{blue}{0.354} & \textcolor{red}{0.177} \\
Lung Lesion                & \textcolor{blue}{0.109} & \textcolor{red}{0.000} & 0.087 \\
Edema                      & \textcolor{blue}{0.046} & \textcolor{red}{0.000} & 0.010 \\
Consolidation              & \textcolor{blue}{0.032} & \textcolor{red}{0.000} & 0.030 \\
Pneumonia                  & \textcolor{red}{0.042} & 0.267 & \textcolor{blue}{0.286} \\
Atelectasis                & 0.135 & \textcolor{red}{0.091} & \textcolor{blue}{0.410} \\
Pneumothorax               & 0.021 & \textcolor{red}{0.000} & \textcolor{blue}{0.040} \\
Pleural Effusion           & \textcolor{red}{0.080} & \textcolor{blue}{0.387} & 0.343 \\
Pleural Other              & \textcolor{blue}{0.011} & 0.005 & \textcolor{red}{0.000} \\
Fracture                   & 0.076 & \textcolor{red}{0.000} & \textcolor{blue}{0.087} \\
Support Devices            & 0.093 & \textcolor{red}{0.000} & \textcolor{blue}{0.100} \\
No Finding                 & \textcolor{red}{0.749} & \textcolor{blue}{0.803} & 0.788 \\
\bottomrule
\end{tabular}
\end{table}

\begin{table}[ht]
\centering
\caption{Per‐domain F1 scores for BiomedCLIP under three settings. `zs', `ft', and `lp' represent zero-shot, fine-tuning, and linear probing, respectively. We highlight the best performance in blue and the worst in red.}
\label{tab:per_domain_f1}
\begin{tabular}{lccc}
\toprule
\textbf{Disease Domain} & \textbf{\textit{zs}} & \textbf{\textit{ft}} & \textbf{\textit{lp}} \\
\midrule
Cardiovascular & \textcolor{red}{0.238} & \textcolor{blue}{0.455} & 0.381 \\
Pulmonary      & \textcolor{blue}{0.453} & 0.387 & \textcolor{red}{0.242} \\
Skeletal       & 0.076 & \textcolor{red}{0.000} & \textcolor{blue}{0.087} \\
Device         & \textcolor{blue}{0.093} & \textcolor{red}{0.000} & 0.076 \\
\bottomrule
\end{tabular}
\end{table}



At the \emph{domain} level (Table \ref{tab:per_domain_f1}), fine-tuning leads overall, but linear probing improves substantially over zero‐shot for Cardiovascular (0.381 vs. 0.238) and Skeletal (0.087 vs. 0.076).  Thus, while full fine‐tuning delivers the best absolute performance, especially on common labels, linear probing offers a highly efficient alternative, which is computationally more feasible.

\subsection{Experiment 3: Qualitative Grad-CAM inspection}

\begin{tcolorbox}[
    colback=blue!10!white,    
    colframe=blue!60!black,   
    boxrule=0.8pt,            
    arc=3pt,                  
    left=6pt, right=6pt,      
    top=3pt, bottom=3pt       
]
\textbf{Key takeaways:}
\begin{itemize}[leftmargin=*]
  \item Zero-shot BiomedCLIP generates Grad-CAM heatmaps that align very closely with radiologist-annotated regions, demonstrating that its visual encoder already encodes rich `where' information for most diseases without any in-domain tuning.

  \item  Fine-tuning produces abstract, non-specific heatmaps that frequently cover irrelevant lung areas.

  \item Linear probing retains nearly all of zero-shot’s spatial fidelity while delivering measurable accuracy gains and generates Grad-CAM heatmap delineates both regions almost as precisely as zero-shot.

  \item Shallower blocks of the model recover compact, ROI-aligned activations; intermediate blocks overgeneralize across the lungs; and the deepest block produces only sparse representations, often missing large lesion areas altogether. 

\end{itemize}
\end{tcolorbox}
In Figure \ref{fig:gradcamsample}, we present two of the fifteen samples annotated by a radiologist. We present the Grad-CAM visualizations generated from zero-shot, fine-tuning, and linear probing of the same radiographs. Additionally, we investigate the information representation across the last three odd layers.

\noindent \textbf{Comparing the explainability BiomedCLIP.} The Grad‐CAM analyses reveal that, in zero‐shot, BiomedCLIP exhibits robust spatial priors for thoracic pathology: confidence values in the range of approximately $\approx0.65$–$0.70$ and heatmaps that are tightly co‐localized with radiologist‐annotated regions, whether for focal lung lesions or combined atelectasis and pleural effusion. These results indicate that the pretrained model’s visual encoder inherently encodes ``\textit{where}'' information for a variety of chest abnormalities, without any in‐domain parameter updates. 
Heatmaps from the fine-tuning often span irrelevant lung fields, and confidence values drop to approximately $0.47$–$0.50$. By contrast, linear probing yields intermediate accuracy improvements over zero‐shot while preserving nearly all of the pretrained spatial fidelity. In the second radiologist-annotated sample, for instance, the linear probe heatmaps at fifth last layer delineate both the collapsed lower‐lobe region and the effusion interface with comparable precision to zero‐shot, whereas fine‐tuning produces a broad, indistinct activation pattern.

\noindent \textbf{Examining the block depths of BiomedCLIP.} A deeper examination of block depth underscores that earlier convolutional stages retain the most interpretable “where” information after transfer. In the fine‐tuned model, activations from the final block (layer -1) are restricted to sparse “pin‐pricks,” often omitting large lesion areas; intermediate blocks (layer -3) generate overly uniform saliency across the lungs; but activations from an earlier block (layer -5) recover compact clusters that align closely with ground‐truth ROIs. This hierarchy suggests that shallow filters capture spatial localization more robustly, whereas deeper filters become overly specialized to classification when trained on limited data.


\section{Conclusion}
\label{concl}

In this work, we presented a comprehensive evaluation of BiomedCLIP \---\ a large, pretrained biomedical vision–language model \---\ on a highly imbalanced, out-of-distribution radiology dataset (IU-Xray).  We compared three adaptation regimes: zero-shot inference, linear probing, and full fine-tuning.  Quantitatively, full fine-tuning achieved the highest inter-/intra-class embedding separability (ratio = 1.78) and the best overall classification metrics (Macro-F1 = 0.235; exact-match = 13.4\%; LRAP = 0.779).  Linear probing recovered much of these gains (Macro-F1 = 0.183; exact-match = 4.3\%; LRAP = 0.741) at a small fraction of the computational cost, while zero-shot inference suffered from high false-positive rates on rare labels (Macro-F1 = 0.105; exact-match = 0.0\%; LRAP = 0.250).

Our embedding-space analysis revealed that end-to-end fine-tuning tightens intra-class clusters more than it increases inter-class distances, whereas linear probing preserves the pretrained representation.  On a per-class level, both adaptation strategies struggled on extremely rare diseases (e.g., Pneumothorax, Consolidation, Edema), but linear probes outperformed full fine-tuning on several mid-frequency conditions (e.g., Pneumonia, Atelectasis, Fracture).  

Qualitative Grad-CAM inspections against 15 radiologist annotated samples further showed that zero-shot BiomedCLIP already encodes accurate “where” information, producing tightly localized heatmaps.  Full fine-tuning tended to generate broader, less specific activations, whereas linear probing retained the strong spatial fidelity of zero-shot while highlighting relevant pathologies more reliably.  We also found that shallower encoder layers often yield the most clinically meaningful saliency maps.

Despite these promising results, our study has several limitations.  First, IU-Xray is one modality with automated labels from CheXbert, limiting ground-truth fidelity and diversity.  Second, our qualitative evaluation relies on a small set of radiologist annotations, which may not generalize across institutions or imaging devices.  Third, we evaluated only a single VLM and three adaptation strategies; other foundation models and parameter-efficient tuning methods (e.g., adapters, prompt learning) remain unexplored.

In the future work, we will extend this evaluation to additional biomedical datasets and modalities (e.g., histopathology, dermatology, neuroimaging), incorporating manually curated labels and larger radiologist-in-the-loop studies.  We will investigate class imbalance mitigation techniques (loss reweighting, synthetic augmentation) to boost rare-disease detection, and explore lightweight adaptation methods—such as soft prompts and adapter modules—to balance accuracy, interpretability, and resource efficiency.  Finally, we aim to integrate calibrated uncertainty estimates and deployable model checkpoints into a clinical decision-support pipeline, moving VLMs one step closer to safe, trusted use in real-world healthcare settings.  

\bibliographystyle{acm}
\bibliography{root}
\end{document}